\pdfoutput=1

\documentclass[11pt]{article}
\usepackage{authblk}
\usepackage[]{EMNLP2022}

\usepackage{times}
\usepackage{booktabs}
\usepackage{multirow}
\usepackage{latexsym}
\usepackage{mathtools}
\usepackage{tikz}
\usetikzlibrary{positioning,calc,arrows.meta}

\usepackage[T1]{fontenc}
\usepackage[utf8]{inputenc}

\title{Learning Better Intent Representations for Financial Open Intent
Classification}
\author[1,2]{Xianzhi Li}
\author[1,2]{Will Aitken}
\author[1,2]{Xiaodan Zhu}
\author[3]{Stephen W. Thomas}
\affil[1]{Department of Electrical and Computer Engineering, Queen's University} 
\affil[2]{Ingenuity Labs Research Institute, Queen's University}
\affil[3]{Smith School of Business, Queen’s University \authorcr 
\{21xl17, will.aitken, xiaodan.zhu, stephen.thomas\}@queensu.ca}

\begin{document}
  \maketitle

\begin{abstract}
    With the recent surge of NLP technologies in the financial domain, banks and other
    financial entities have adopted virtual agents (VA) to assist customers. A
    challenging problem for VAs in this domain is determining a user's reason
    or intent for contacting the VA, especially when the intent was unseen or
    \textit{open} during the VA's training. One method for handling open
    intents is adaptive decision boundary (ADB) post-processing, which learns
    tight decision boundaries from intent representations to separate known and
    open intents.  We propose incorporating two methods for supervised
    pre-training of intent representations: prefix-tuning and fine-tuning
    just the last layer of a large language model (LLM). With this proposal,
    our accuracy is 1.63\% - 2.07\% higher than the prior
    state-of-the-art ADB method for open intent classification on the
    banking77 benchmark amongst others.  Notably, we only supplement the original
    ADB model with 0.1\% additional trainable parameters.  Ablation studies
    also determine that our method yields better results than full fine-tuning
    the entire model. We hypothesize that our findings could stimulate a new
    optimal method of downstream tuning that combines parameter efficient
    tuning modules with fine-tuning a subset of the base model's layers.
\end{abstract}

\section{Introduction} 

\setlength{\tabcolsep}{10pt}
\begin{table}
    \centering
    \resizebox{\columnwidth}{!}{%
    \begin{tabular}{lc}
	\toprule
	\textbf{Utterance} & \textbf{Label}\\
	\midrule
	When will I get my card? & Card Arrival\\
	What exchange rates do you offer? & Exchange Rate \\
	My card hasn't arrived yet. & Card Arrival \\
	Is it a good time to exchange? & Exchange Rates\\
	... & ... \\
	Is it possible to get a refund? & \textbf{Open} \\
	Why has my withdrawal not posted? & \textbf{Open} \\
	\bottomrule

    \end{tabular}}
    \caption{Example user utterances and associated intent labels from
	banking77 dataset \citep{Casanueva:20}. In this example, only Card Arrival
	and Exchange Rate intents were known in training and thus refund and withdrawal
	related requests are Open intents in this context.}
    \label{tab:open_ex}
\end{table}

As the popularity of virtual agent (VA) dialogue systems increases and their
application in the finance domain is explored, the problem of intent classification
demands greater attention. Several recent finance-specific VAs leverage
technical advancements to respond to natural language queries
\citep{galitsky-ilvovsky-2019-chatbot, Khan:20}. Determining the user's intent
ensures that the VA can appropriately tailor its responses and/or perform
relevant actions. Initial works in intent classification limited the task to
classifying utterances as one of $N$ known intents and achieved high accuracy
\citep{Weld:21}.  However, as depicted in Table \ref{tab:open_ex}, real-world
applications often encounter intents unseen in training data that can be
considered as \textit{open} in the current context. Accounting for the open
class establishes an $(N+1)$-class classification task
\citep{shu-etal-2017-doc}, where the open class is used as a label for any
unidentified intent. 

An optimal classifier for this problem must balance correctly labelling
known-class utterances while avoiding mistakenly classifying open utterances as
one of the known classes. \citep{Zhang:21} addresses this problem by proposing
a novel loss function to learn an adaptive decision boundary (ADB) for each
known intent. At inference, samples that do not fall within any ADB are
classified as open. Compact intent representations are required as input for
the ADB post-processing learning step and in the case of \citep{Zhang:21} the
representations are learnt by fine-tuning the last layer of BERT
\citep{devlin-etal-2019-bert}. Since most intent classification methods require
post-processing on intent representations, our work focuses on deriving richer
representations by leveraging large language models (LLM) in an efficacious
manner while still minimizing trainable parameters.

Following the introduction of the transformer in \citep{Vaswani:17}, an influx
of LLM architectures have continually progressed state-of-the-art (SOTA)
performance on many natural language processing (NLP) tasks \citep{Otter:21}.
Usually these models are pre-trained on a general self-supervised learning
task, after which they are fine-tuned for a specific task. Fine-tuning such a
model can be computationally prohibitive due to the immense number of trainable
parameters. Furthermore, \citep{Kaplan:20} found that the most important factor
for LLM performance is likely model size, indicating that development of even 
larger models is probable. Inspired by in-context prompting,
\citep{li-liang-2021-prefix} proposed prefix tuning as a parameter efficient
alternative to fine-tuning for natural language generation (NLG). The LLM's
parameters are frozen and trainable prefix tokens are prepended to the
input sequence.  Prefix-tuning has been adapted to natural language
understanding (NLU) and performs comparably to full fine-tuning
across scales and tasks \citep{liu-etal-2022-p}. 

We achieve SOTA results by augmenting the pre-training architecture of ADB open
intent classification \citep{Zhang:21} with prefix-tuning.  The combination of
prefix-tuning with fine-tuning only the last transformer layer was motivated by
\citep{Kumar:22}, which discovered that fine-tuning the entire model can
distort pre-trained features.  We find that alone, both prefix-tuning or
fine-tuning the last layer under-performs fine-tuning all of BERT but when
trained in tandem, exceeds full fine-tuning.

The rest of this paper is structured as follows: Section \ref{sec:rw}
summarizes prior works in both intent classification and parameter efficient
tuning (PET). Our methodology and model architecture are defined in Section
\ref{sec:meth}. In Sections \ref{sec:exp} and \ref{sec:results} respectively,
we provide our experimentation structure and corresponding results as well as
several ablations. We finish with a conclusion and brief discussion regarding
limitations and ethics.

\section{Related Works} \label{sec:rw}
\subsection{Financial Virtual Agents}
The effectiveness of VAs has led to their adoption in the financial domain.
\citep{galitsky-ilvovsky-2019-chatbot} demonstrated an exemplary session with a
financial VA where the user queried for investment advice. CalFE leverages
commercial chatbot frameworks to train a finance-specific VA \citep{Khan:20}.
\citep{Ng:20} evaluates the impact of a VA's social presence on usage
intention in VAs for finance. All of these works require extracting intent
from user utterances.

\subsection{Intent Detection}
Intent classification is a well-established NLU task but most research limits the
problem to known classes \citep{zhang-etal-2019-joint,e-etal-2019-novel,
qin-etal-2019-stack, zhang-etal-2021-shot}. While having prior knowledge of all
expected intents is ideal, this is rarely possible in a production environment,
especially for new dialogue systems. More realistically, a subset of intents
are anticipated and new intents are discovered after deployment.
\citep{brychcin-kral-2017-unsupervised} recognized the challenge of identifying
intents prior to training and proposed an unsupervised method to group intents,
but by doing so, likely ignored information available in the already identified
intents.  \citep{xia-etal-2018-zero} employed zero-shot learning to identify
emerging intents but used an LSTM which is hindered by non-parallelized
learning and challenges in propagating long-range dependencies. The same issue
is present in DeepUnk, a BiLSTM-based intent classification method using margin loss
\citep{lin-xu-2019-deep}. \citep{zhan-etal-2021-scope} shared our open intent
classification problem formulation but synthetically generated out-of-domain samples
for training which may not be as realistic as a fine-grained open class
representation. 

Our work directly extends the ADB approach to establishing an open class
representation \citep{Zhang:21}. The novelty of our adaptation is in leveraging
prefix tuning in combination with partial fine-tuning to improve the
pre-training of known intent representations without drastically increasing the
number of trainable parameters. In parallel with our work, \citep{Zhang:22}
extended their ADB approach to learn distance-aware intent representations.
Doing so resulted in comparable performance to our modification of their
original approach. However, our tuning method is model-agnostic and can easily
be incorporated with their distance-aware representation learning, likely
improving the SOTA further.

\subsection{Parameter Efficient Tuning}

The desire for PET quickly emerged following the introduction of LLMs. Adapter
modules insert task-specific parameters sequentially between transformer layers
while the rest of the model remains frozen \citep{Houlsby:19}. 
\citep{li-liang-2021-prefix} and \citep{lester-etal-2021-power} simultaneously
substantiated the efficacy of prepending tokens to attention mechanisms as a
means of efficient tuning. In \citep{li-liang-2021-prefix}, the prefixes are
applied at each layer of the transformer while \citep{lester-etal-2021-power}
only prepends to the input sequence. \citep{liu-etal-2022-p} applied the same
method to NLU tasks using deep prefixes with optional reparameterization.
Without reparameterization, simple embeddings are learnt for the prefixes.
Reparameterization inserts a multi-layer perceptron (MLP) between the
embeddings and prefix tokens which allows for more complex embeddings. 

Recently, \citep{He:22} determined the theoretical impact
of various PET methods and deduced that they are all modifications of a similar
function. Allocating additional parameters to other PET modules as suggested 
by \citep{He:22} could optimize intent representation beyond what is possible
with prefixes alone. For now we limit our work to the most efficient method for
low resource settings, prefix-tuning. To the best of our knowledge, this is the
first PET work to combine partial fine-tuning with prefix-tuning.

\section{Methodology} \label{sec:meth}
In this section we explain our procedure for open intent classification.
Section \ref{sec:pt} describes prefix-tuning, the method we supplement partial
fine-tuning with. Section \ref{sec:training} provides a brief summary of
training the original ADB method that we have extended \citep{Zhang:21}.

\subsection{Prefix-Tuning} \label{sec:pt}
Prefix-tuning prepends trainable prefix tokens $P_{k}$,$P_{v}$ in front of
Key and Value vectors of multi-head attention in each transformer layer. The
attention mechanism is applied to the concatenation of prefix and original
tokens. Equation \ref{eq:pt_att} details the computation.

\begin{equation} \label{eq:pt_att}
    \begin{split}
	head = Soft&max(Q * Concat(P_k,K)^T) \\
	     & * Concat(P_{v},V)
    \end{split}
\end{equation}
Where Q, K, and V are the Query, Key and Value matrices from the original
transformer \citep{Attention:17}. $P_{k}$ and $P_{v}$ are the additional prefix 
tokens and are prepended to the Key and Value matrices.

Often, prefix-tuning methods use a MLP to
reparameterize the prefix since directly embedding can lead to
unstable training and performance decrease \citep{li-liang-2021-prefix}.
However, \citep{liu-etal-2022-p} found that for NLU tasks, the efficacy of
reparameterization is dependent on the task.  From our experiments, we
determine that reparameterizing the prefixes is crucial for intent classification.
Following training, the MLP weights and biases from reparameterization are 
dropped and only prefixes are kept. 

\begin{figure}
    \centering
    \begin{tikzpicture}
	[tuned/.style={rectangle, draw=orange!50, fill=orange!20, thick,
		       inner sep=1.5mm, minimum height=7mm, minimum width=1.1cm, 
		       rounded corners=1mm},
	 fixed/.style={rectangle, draw=blue!50, fill=blue!20, thick,
		       inner sep=1.5mm, minimum height=7mm, minimum width=1.6cm, 
		       rounded corners=1mm},
	 transformer_fixed/.style={rectangle, draw=blue!50, fill=blue!20, thick,
		       inner sep=1.5mm, minimum height=7mm, minimum width=11.45cm, 
		       rounded corners=1mm},
	 transformer_tuned/.style={rectangle, draw=orange!50, fill=orange!20, thick,
		       inner sep=1.5mm, minimum height=7mm, minimum width=11.45cm, 
		       rounded corners=1mm},
	 scale=0.67, transform shape]

        \node (p_1_1) [tuned] {$P_{1,1}$};
	\node (p_1_2) [tuned, right = 3mm of p_1_1] {$P_{1,2}$};
	\node (f_1_0) [fixed, right = 3mm of p_1_2] {[$CLS$]};
	\node (f_1_1) [fixed, right = 3mm of f_1_0] {$Tok_1$};
	\node (f_1_2) [fixed, right = 3mm of f_1_1] {$Tok_2$};
	\node (f_1_n) [fixed, right = 1.5cm of f_1_2] {$Tok_n$};
	\node at ($(f_1_2)!.5!(f_1_n)$) {\ldots};

	\node (t_1) [transformer_fixed, above = 3mm of p_1_1.north west, 
	    anchor=south west] {Transformer Layer 1};

	\foreach \emb in {p_1_1, p_1_2, f_1_0, f_1_1, f_1_2, f_1_n}
	    \draw [black!50] (\emb.north) -- +(0,0.3);

	\node (p_2_1) [tuned, above = 5mm of t_1.north west, 
	    anchor=south west] {$P_{2,1}$};
	\node (p_2_2) [tuned, right = 3mm of p_2_1] {$P_{2,2}$};
	\node (f_2_0) [fixed, right = 3mm of p_2_2] {};
	\node (f_2_1) [fixed, right = 3mm of f_2_0] {};
	\node (f_2_2) [fixed, right = 3mm of f_2_1] {};
	\node (f_2_n) [fixed, right = 1.5cm of f_2_2] {};

	\foreach \emb in {f_2_0, f_2_1, f_2_2, f_2_n}
	    \draw [black!50] (\emb.south) -- +(0,-0.5);

	\node (t_2) [transformer_fixed, above = 3mm of p_2_1.north west, 
	    anchor=south west] {Transformer Layer 2};
	\node at ($(f_2_2)!.5!(f_2_n)$) {\ldots};

	\foreach \emb in {p_2_1, p_2_2, f_2_0, f_2_1, f_2_2, f_2_n}
	    \draw [black!50] (\emb.north) -- +(0,0.3);

	\node (p_3_1) [tuned, above = 1cm of t_2.north west, 
	    anchor=south west] {$P_{N,1}$};
	\node (p_3_2) [tuned, right = 3mm of p_3_1] {$P_{N,2}$};
	\node (f_3_0) [fixed, right = 3mm of p_3_2] {};
	\node (f_3_1) [fixed, right = 3mm of f_3_0] {};
	\node (f_3_2) [fixed, right = 3mm of f_3_1] {};
	\node (f_3_n) [fixed, right = 1.5cm of f_3_2] {};
	\node at ($(f_3_2)!.5!(f_3_n)$) {\ldots};

	\foreach \emb in {f_3_0, f_3_1, f_3_2, f_3_n}
	    \draw [loosely dash dot dot, black!50] (\emb.south) -- +(0,-1);

	\node (t_3) [transformer_tuned, above = 3mm of p_3_1.north west, 
	    anchor=south west] {Transformer Layer N};

	\foreach \emb in {p_3_1, p_3_2, f_3_0, f_3_1, f_3_2, f_3_n}
	    \draw [black!50] (\emb.north) -- +(0,0.3);

	\node (f_4_0) [fixed, above = 1.4cm of f_3_0.north west,
	    anchor=south west] {};
	\node (f_4_1) [fixed, right = 3mm of f_4_0] {};
	\node (f_4_2) [fixed, right = 3mm of f_4_1] {};
	\node (f_4_n) [fixed, right = 1.5cm of f_4_2] {};

	\node at ($(f_4_2)!.5!(f_4_n)$) {\ldots};

	\node (mean_pool) [fixed, above = 1.5cm of t_3.north] {Mean Pooling};
	\node (dense) [tuned, above = 0.4cm of mean_pool.north] {Dense Layer};

	\foreach \emb in {f_4_0, f_4_1, f_4_2, f_4_n} {
	    \draw [black!50] (\emb.south) -- +(0,-0.4);
	    \draw [black!50] (\emb.north) -- (mean_pool.south);
	}

	\draw [black!50] (mean_pool) -- (dense);
	\node (intents) [fixed, above = 0.4cm of dense.north] 
	    {Intent Representation};
	\draw [black!50] (dense) -- (intents);
	\node (lc) [tuned, above = 0.4cm of intents.north] 
	    {Linear Classifier};
	\draw [black!50] (lc) -- (intents);

    \end{tikzpicture}
    \caption{Pre-training architecture for learning intent representations.
	Orange blocks denote trainable parameters while blue are fixed. For
	this concrete example, the prefix length has been set to two, but this
	value is a tunable hyperparameter.}
    \label{fig:arch}
\end{figure}
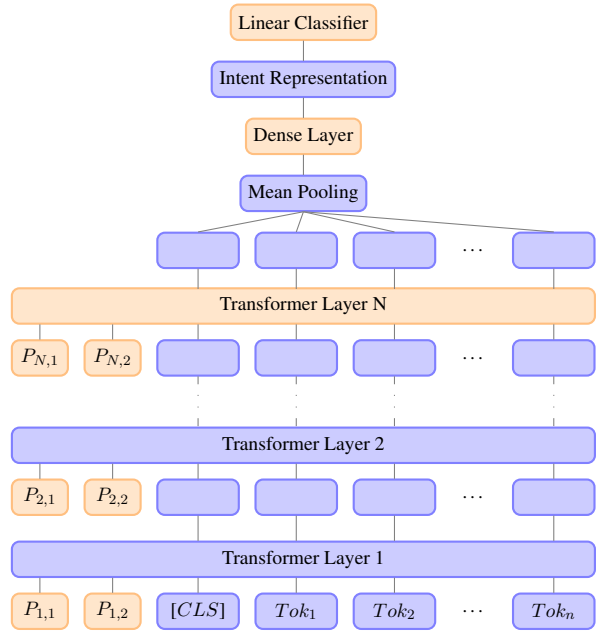

\setlength{\tabcolsep}{12pt}
\begin{table*}
    \centering
    \resizebox{.95\textwidth}{!}{%
    \begin{tabular}{llcccccc}
	\toprule

	& & \multicolumn{2}{c}{BANKING} & \multicolumn{2}{c}{OOS} 
	    & \multicolumn{2}{c}{StackOverflow} \\
	\cmidrule(rl){3-4} \cmidrule(rl){5-6} \cmidrule(rl){7-8}
	KIR & Method & Accuracy & F1-Score & Accuracy & F1-Score & 
	    Accuracy & F1-Score \\

	\midrule

	\multirow{4}*{25\%} & DeepUnk & 64.21 & 61.36 & 81.43 & 71.16 &
	    47.84 & 52.05 \\
	& $(K+1)$-way & 74.11 & 69.93 & / & / & 68.74 & 65.64 \\
	& ADB & 78.85 & 71.62 & 87.59 & 77.19 & 86.72 & \textbf{80.83} \\
	& PFT-ADB & \textbf{80.14} & \textbf{72.86} & \textbf{88.03} &
	    \textbf{78.85} & \textbf{87.60} & 80.78 \\

	\midrule

	\multirow{4}*{50\%} & DeepUnk & 72.73 & 77.53 & 83.35 & 82.16 &
	    58.98 & 68.01 \\
	& $(K+1)$-way & 72.69 & 79.21 & / & / & 75.08 & 78.55 \\
	& ADB & 78.86 & 80.90 & 86.54 & 85.05 & 86.40 & 85.83 \\
	& PFT-ADB & \textbf{80.40} & \textbf{82.44} & \textbf{87.60} &
	    \textbf{86.87} & \textbf{87.06} & \textbf{86.22} \\

	\midrule

	\multirow{4}*{75\%} & DeepUnk & 78.52 & 84.31 & 83.71 & 86.23 &
	    72.33 & 78.28 \\
	& $(K+1)$-way & 81.07 & 86.98 & / & / & 81.71 & 85.85 \\
	& ADB & 81.08 & 85.96 & 86.32 & 88.53 & 82.78 & 85.99 \\
	& PFT-ADB & \textbf{82.76} & \textbf{87.35} & \textbf{88.94} &
	    \textbf{90.93} & \textbf{83.46} & \textbf{86.61} \\

	\bottomrule
    \end{tabular}}
    \caption{Main results for known intent ratios (KIR) 25\%, 50\%, and 75\% on
	BANKING, OOS, and StackOverflow datasets. Average accuracy and macro F1-Score 
	are reported over all classes.}
    \label{tab:main_results}
\end{table*}

\subsection{Training} \label{sec:training}
Figure \ref{fig:arch} illustrates our pre-training architecture of 
prefix-tuning plus tuning the last transformer layer to extract intent
representations. The orange components of the diagram are trainable and the
blue are frozen. This example shows a prefix length of two, but the length is
a flexible hyperparameter. We detail our entire hyperparameter settings in 
Section \ref{sec:settings}. The outputs of BERT are first fed into a
mean-pooling function to aggregate the sequence into a single vector $x_i$ as
described by Equation \ref{eq:mp}:

\begin{equation} \label{eq:mp}
    \begin{split}
	x_{i} = mp(& [CLS], Tok_1,Tok_2,...,Tok_{M})
    \end{split}
\end{equation}
\noindent
where $i$ refers to the current training sample. A dense layer transforms the
vector to the intent representation feature space and the resultant vector is
finally passed to a linear classifier. We pre-train on known intents and their
labels with softmax as the loss function to optimize both the prefix
tokens and the last transformer layer. Equation \ref{eq:loss} is the softmax loss:

\begin{equation} \label{eq:loss}
	Loss=-\frac{1}{n}\sum_{c=1}^n\log(\frac{e^{z_{i}}}{\sum_{j=1}^K e^{z_{j}}})
\end{equation}
\noindent
where $n$ is the batch size and $z_{j}$ refers to the output logits of $j_{th}$
class. 

Following pre-training, the intent representations are extracted from
our model for ADB post-processing. ADB learns a tight
spherical decision boundary for each known intent. At inference, intent
representations that fall outside of all decision boundaries are classified as
open. For clarification, the only alteration to the ADB method we employ is the
addition of prefix tokens in Figure \ref{fig:arch}. See \citep{Zhang:21} for more
information regarding decision boundaries and other training details.

\section{Experiments} \label{sec:exp}
\subsection{Datasets}
\textbf{BANKING}:
A dataset of 77 banking intents with samples summing to 13,083 
banking-specific customer service queries \citep{Casanueva:20}. It
is also commonly referred to as ``banking77'' but \citep{Zhang:21} uses
``BANKING'' and since we are comparing our results primarily with them, we 
conform 
to their choice. \newline  \newline
\textbf{OOS}:
A subset of CLINC50 specifically designed for out-of-scope intent prediction
\citep{larson-etal-2019-evaluation} with 22,500 and 1,200 in and out of domain
samples respectively over 150 different intents spanning 10 domains. \newline
\newline
\textbf{StackOverflow}:
The processed version of the StackOverflow dataset \citep{xu-etal-2015-short},
which has 20 different intents and 1,000 samples for each. 

\begin{table*}
    \centering
    \resizebox{.82\textwidth}{!}{%
    \begin{tabular}{llcccccc}
	\toprule

	& & \multicolumn{2}{c}{BANKING} & \multicolumn{2}{c}{OOS} 
	    & \multicolumn{2}{c}{StackOverflow} \\
	\cmidrule(rl){3-4} \cmidrule(rl){5-6} \cmidrule(rl){7-8}
	KIR & Method & Open & Known & Open & Known & Open & Known \\

	\midrule

	\multirow{4}*{25\%} & DeepUnk & 70.44 & 60.88 & 87.33 & 70.73 &
	    49.29 & 52.60 \\
	& $(K+1)$-way & 80.12 & 69.39 & / & / & 74.86 & 63.80 \\
	& ADB & 84.56 & 70.94 & 91.84 & 76.80 & 90.88 & \textbf{78.82} \\
	& PFT-ADB & \textbf{85.65} & \textbf{72.19} & \textbf{92.11} &
	    \textbf{78.50} & \textbf{91.58} & 78.62 \\

	\midrule

	\multirow{4}*{50\%} & DeepUnk & 69.53 & 77.74 & 85.85 & 82.11 &
	    43.01 & 70.51 \\
	& $(K+1)$-way & 67.26 & 79.52 & / & / & 71.88 & 79.22 \\
	& ADB & 78.44 & 80.96 & 88.65 & 85.00 & 87.34 & 85.68 \\
	& PFT-ADB & \textbf{80.02} & \textbf{82.51} & \textbf{89.34} &
	    \textbf{86.83} & \textbf{88.17} & \textbf{86.02} \\

	\midrule

	\multirow{4}*{75\%} & DeepUnk & 58.54 & 84.75 & 81.15 & 86.27 &
	    37.59 & 81.00 \\
	& $(K+1)$-way & 60.71 & 87.47 & / & / & 65.44 & 87.22 \\
	& ADB & 66.47 & 86.29 & 83.92 & 88.58 & 73.86 & 86.80 \\
	& PFT-ADB & \textbf{69.18} & \textbf{87.66} & \textbf{86.80} &
	    \textbf{90.96} & \textbf{74.78} & \textbf{87.40} \\

	\bottomrule
    \end{tabular}}
    \caption{Open and known comparison of main results for known intent 
	ratios 25\%, 50\%, and 75\% on BANKING, OOS, and StackOverflow 
	datasets. F1-Score and macro F1-Score are reported for open class and
	known classes respectively.}
    \label{tab:main_open_known}
\end{table*}

\subsection{Experiment Settings} \label{sec:settings}
In accordance with previous methods, we sample 25\%, 50\%, and 75\% of intent
classes randomly during training as the ``known'' classes. The remaining are set
aside as open classes and removed from training sets. We use BERT
(bert-base-uncased) provided by Hugging Face \citep{wolf-etal-2020-transformers}
to extract intent representations from utterances. The learning rate for
prefixes and transformer parameters is set to 2e-5 since experimenting with setting
different learning rates for prefixes and last layer of transformer did not
consistently lead to a performance increase.  All experiments are conducted on
a NVIDIA 2080TI GPU. To fairly compare our method, we keep other
hyperparameters the same as \citep{Zhang:21}.  For all results we average
performance over ten random seeds.

Regarding prefix-specific settings, we use reparameterization with a hidden
size of 512 unless otherwise specified. The overall parameter size is determined by the prefix length. In this task, we found that enlarging the prefix length did
not lead to a consistent performance increase due to its low-rank bottleneck.
\citep{He:22} also discusses that allocating additional parameter in
self-attention is only worthwhile if they make up less than 0.1\% of the
parameter budget. Therefore, we choose our default prefix length as 10, which
equates to roughly 0.1\% of BERT's trainable parameters.

\subsection{Baselines}

We compare our results to the most competitive open intent classification
methods: DeepUnk \citep{lin-xu-2019-deep}, $(K+1)$-way
\citep{zhan-etal-2021-scope}, and the ADB method we directly extend
\citep{Zhang:21}. The DeepUnk results are taken from \citep{Zhang:21} which
replaced the BiLSTM with BERT to generate intent representations for fair
comparison. \citep{zhan-etal-2021-scope} also uses BERT as its encoder but
keeps just the CLS token's final hidden state instead of pooling the entire
sequence. \citep{zhan-etal-2021-scope} did not test on the same OOS split and
cells corresponding to that configuration are left blank for tables in Section
\ref{sec:results}.

\section{Results} \label{sec:results}

Our main results and respective baseline comparisons are presented in Tables
\ref{tab:main_results} and \ref{tab:main_open_known}. Table
\ref{tab:main_results} is limited to
accuracy averaged over all classes, including the open class and macro F1
over the same set of classes. For a fine-grained analysis of open intent
performance, Table \ref{tab:main_open_known} contrasts the F1 score of the open
class with the macro F1 over the remaining known classes.
\textit{PFT-ADB} denotes our method of adding prefix tuning to ADB and 
the best result for each section is in boldface.

For each dataset we tested, PFT-ADB improves performance on all prior methods
with the minor exception of StackOverflow F1-Score and known score.
Specifically, as shown in Table \ref{tab:main_results}, we achieve accuracy
improvements of (1.63\%, 1.95\%, 2.07\%) on BANKING, (0.50\%, 1.22\%, 3.03\%)
on OOS, and (1.01\%, 0.76\%, 0.82\%) on StackOverflow for known intent ratios
(25\%, 50\%, 75\%). The consistency of our results across configurations
suggests that paying closer attention to pre-training intent representations
can enhance the distinction of decision boundaries in the post-processing step.
Additionally, we do not add a significant number of trainable parameters to
existing methods (only 0.1\%), successfully avoiding trading substantial costs
for performance increase. Note that our results are comparable to that of the
most recently released DA-ADB \citep{Zhang:22} model. We believe that due to
their orthogonal nature, DA-ADB and our approach could be combined together for
further performance improvements.

We note that the dataset with the lowest performance gain is StackOverflow.
\citep{Zhang:21} found that their novel post-processing method, ADB, was most
effective on this dataset compared to prior methods. They hypothesized that
this was due to being able to form tighter decision boundaries for the
technical jargon more prevalent in StackOverflow.  Following this reasoning, it
could be that for this dataset the post-processing method is paramount and
enriching the intent representations alone is not enough to yield a substantial
performance improvement.

It is important that an open intent classification method balances the performance
on known classes while still identifying open intents. Table
\ref{tab:main_open_known} verifies that despite changing pre-training tuning
methods, ADB post-processing still adequately addresses this issue. The
performance increase is consistent between both the open class and known
classes for each dataset indicating that prefix-tuning does not interfere with
optimizing both aspects of the open intent problem. Again, we anticipate that
combining PFT-ADB with the newer DA-ADB could result in even better performance.

The following ablations focus on the OOS dataset since it covers multiple
domains and we wanted to generalize beyond just the financial domain.

\subsection{Effect of Reparameterization and Tuning Variations}

\setlength{\tabcolsep}{6pt}
\begin{table}
    \resizebox{\columnwidth}{!}{%
    \begin{tabular}{lcccc} 
	\toprule
	Method & Accuracy & F1-score & Open & Known \\
	\midrule

	Emb	   & 64.40 & 68.56 & 66.62 & 68.38\\
	MLP	   & 81.56 & 85.89 & 75.98 & 85.97\\
	Emb+12th L & 86.40 & 88.15 & 84.44 & 88.19\\
	MLP+12th L & \textbf{90.07} & \textbf{91.52} & \textbf{88.48} &
	\textbf{91.54} \\
	FFT-NoPT   & 87.56 & 89.20 & 85.71 & 89.24 \\
	ADB	   & 86.32 & 88.53 & 83.92 & 88.58 \\

	\bottomrule
    \end{tabular}}
    \caption{Experiments on the impact of different prefix encoding approaches
	with and without fine-tuning the last layer of transformer.  We use OOS
	dataset with 75\% known Intent Ratio.  ``Emb'' refers to embedding-only
	method. ``MLP'' refers to method that uses 2 layers of MLP to encode prefix.  
	``+12th L'' means we unfreeze the last layer of transformer. ``FFT-NoPT''
	denotes full fine-tuning without any prefixes.}
    \label{tab:reparam}
\end{table}

In Table \ref{tab:reparam} we show that under the same dataset and known intent
ratio, performance varies considerably when adopting MLP as prefix encoder. In
the first row, the embedding-only method leads to poor results of 64.40\%
accuracy. Contrarily, introducing a 2 layer MLP to encode prefixes increases
the performance by around 15\%.  More importantly, the result is stable and
reproducible. It indicates that using MLP to reparameterize prefixes is crucial
in obtaining a consistent performance. 

Results using prefix tuning alone (rows 1 and 2) in this task are
slightly worse than ADB’s fine-tuning results. In particular, the performance
gap in identifying open intent is more salient,
revealing prefix-tuning's lower capacity for out-of-scope
classification. However, when we incorporate prefix tuning along with tuning
the last layer of transformer, we find a surprisingly large performance
increase. For embedding and MLP methods, tuning the last layer of transformer
gives a performance boost to 86.40\% and 90.07\%, respectively, with only
additional 0.1\% of ADB's parameters. Since the latter transformer layer
captures high-level feature of utterances, we believe that this small amount of
parameter steer the higher layers to learn more task-oriented
information as well as fit intents into a better-distributed latent space. 

We also try the common method of fully fine-tuning, i.e., unfreezing all of
BERT's parameters which was not done in \citep{Zhang:21}. The performance is still 1\% lower than our method while we use only 8.1\% of parameters.

\subsection{Impact of Prefix Lengths}

\setlength{\tabcolsep}{6pt}
\begin{table}
    \resizebox{\columnwidth}{!}{%
    \begin{tabular}{lccccc} 
	\toprule
	Method & Length & Accuracy & F1-score & Open & Known \\
	\midrule

	\multirow{6}*{Emb+12th L} &
	  10  & 87.60 & 89.08 & 85.82 & 89.11 \\
        & 20  & \textbf{88.25} & \textbf{89.49} & \textbf{86.80} &
	    \textbf{89.52} \\
	& 30  & 87.88 & 89.44 & 86.13 & 89.47 \\
	& 50  & 85.70 & 87.49 & 83.72 & 87.53 \\
	& 80  & 85.93 & 87.75 & 83.87 & 87.78 \\
	& 100 & 87.95 & 89.45 & 86.16 & 89.48 \\

	\midrule
	\multirow{6}*{MLP+12th L} &
	10  & \textbf{90.16} & 91.57 & \textbf{88.57} & 91.60\\
	&  20  & 89.67 & 91.27 & 87.96 & 91.30\\
	&  30  & 90.05 & \textbf{91.59} & 88.35 & \textbf{91.62} \\
	&  50  & 88.65 & 90.34 & 86.82 & 90.37\\
	&  80  & 89.49 & 91.27 & 87.51 & 91.30\\
	&  100 & 89.84 & 91.51 & 88.09 & 91.54\\

	\bottomrule
    \end{tabular}}
    \caption{Results of tuning with different prefix lengths. We use OOS
	dataset with 75\% known intent ratio.}
    \label{tab:length}
\end{table}

We experimented with the prefix length to determine its effect on performance. 
From Table \ref{tab:length}, we observe that with the increase of the prefix length from 10 to 100 (parameter size from .1\% to 1.6\%), the results do not follow the same ascending pattern. We argue that simply adding more prefix tokens
would not lead to a consistent performance boost due to its bottleneck.
\citep{He:22} determined that prefix tuning is another form of low-rank update,
which cannot make use of more than 0.1\% of additional parameters. 

\subsection{Fine-Tuning Different Groupings of Layers}
Combining prefix-tuning with fine-tuning a subset of transformer layers is, to
the best of our knowledge, a novel approach. Fine-tuning the last layer alone is
ideal for minimizing trainable parameters. We aim to determine whether
varying which layer or group of several layers is unfrozen can achieve better
results than the last layer alone. Table \ref{tab:layer_group} summarizes our
findings. The layer of interest is specified with the variable $x$.  ``Just $x$''
is fine-tuning layer $x$ alone and ``$x$ and Rest'' is fine-tuning the layer $x$
and all subsequent layers. Using this notation, ``Layer \textit{1} and Rest'' is 
akin to fine-tuning all of BERT. ``No-FT'' refers to prefix-tuning without any
additional fine-tuning. For this row and when $x$ is 12, the results between
the two main columns are of course the same.

\setlength{\tabcolsep}{8pt}
\begin{table}
    \centering
    \resizebox{\columnwidth}{!}{%
    \begin{tabular}{ccccc}
	\toprule

	\multirow{2}*{$x$} & \multicolumn{2}{c}{Just $x$} 
	    & \multicolumn{2}{c}{$x$ and Rest} \\
	\cmidrule(lr){2-3} \cmidrule(lr){4-5} 
	& Accuracy & F1-Score & Accuracy & F1-Score \\
	\midrule

	No-FT & 81.56 & 85.89 & 81.56 & 85.89 \\
	12 & \textbf{90.07} & \textbf{91.62} & 90.07 & 91.62 \\
	11 & 89.18 & 91.19 & \textbf{90.21} & \textbf{91.81} \\
	10 & 89.81 & 91.80 & 89.42 & 91.42 \\
	9 & 88.81 & 90.89 & 88.93 & 91.18 \\
	7 & 87.07 & 89.79 & 88.42 & 90.93 \\
	4 & 85.23 & 88.21 & 87.00 & 89.90 \\
	1 & 84.77 & 87.73 & 87.58 & 90.24 \\

	\bottomrule
    \end{tabular}}
    \caption{Fine-tuning various groupings of transformer layers on OOS with
	known intent ratio 75\%. ``No-FT'' is prefix-tuning without any
	fine-tuning. Prefix-tuning configuration was kept constant throughout
	runs.}
    \label{tab:layer_group}
\end{table}

Several interesting observations are evident in Table \ref{tab:layer_group}.
Firstly, the fine-tuning of at least one layer in addition to prefix-tuning is
strictly necessary for optimal performance. Under the constraint of tuning just
a single layer, the last performs the best. The latter layers of the model are
where higher-level details of natural language are processed. We hypothesize
that tuning this layer best incorporates the propagation of prior prefixes with
the base model. Tuning prior layers may have a similar effect, but if the
subsequent layers are frozen, the understanding of prompts is obfuscated since
the latter frozen layers have no experience attending to prefixes.

Another notable finding is that if performance is to be prioritized,
fine-tuning the final two layers together is better than the last layer
alone. This suggests that the prefixes are complex enough such that their 
value is maximized when the final two layers tune in tandem. However, the trade
off of minor performance increase at the cost of doubling the trainable
parameters may not be worth it depending on the application.

Lastly, we note that as layers beyond two are trained in the ``$x$ and Rest'' 
column, performance begins to degrade. This supports the observation made by
\citep{Kumar:22} that fine-tuning disturbs pre-trained features in the base
model. Training only the final layer(s) avoids perturbing low-level semantics
learnt in earlier layers of the base model, but still adds sufficient capacity 
to attend to the prefixes.

\setlength{\tabcolsep}{5pt}
\begin{table}
    \centering
    \resizebox{\columnwidth}{!}{%
    \begin{tabular}{lcccc}
	\toprule

	Method & Accuracy & F1-Score & Open & Known \\

	\midrule

	Attention & 87.49 & 89.69 & 85.19 & 89.73 \\
	Feed Forward & 86.47 & 88.92 & 83.81 & 88.96 \\
	Layer Normalization & 86.61 & 88.81 & 84.21 & 88.85 \\
	Keys and Values & 85.67 & 88.58 & 82.39 & 88.64 \\
	Entire Layer & \textbf{90.07} & \textbf{91.52} & \textbf{88.48} &
	    \textbf{91.54} \\

	\bottomrule
    \end{tabular}}
    \caption{Fine-tuning components of final transformer layer on OOS with
	known intent ratio 75\%. Only the parameters of the component(s) are
	tuned and the rest of the layer is frozen.}
    \label{tab:layer_component}
\end{table}

\subsection{Fine-Tuning Various Components in Last Layer}
While fine-tuning only the last transformer layer reduces the trainable
parameter count to 8\%, this is still a large value compared to the 0.1\%
parameter count of the prefixes alone. We isolate various components of the
last transformer layer to determine if some could be frozen to further reduce
parameter count.  The results are presented in Table \ref{tab:layer_component}.
Tuning the entire layer significantly outperformed any other variation, alluding
that there is an important relationship between the prefixes and every
component of the final transformer layer. Tuning each of the components in the
last layer is essential to procure maximum prefix performance.

\section{Conclusion} \label{sec:conc}
We have shown that incorporating prefix-tuning with the ADB intent
representation pre-training method achieves SOTA results in the financial
domain on the banking77 benchmark dataset and others. Furthermore, our tuning
method does not sacrifice excessive parameters count for the performance gain.
The combination of prefix-tuning with fine-tuning only the last layer of
transformer is simple yet novel to the best of our knowledge and surfaces
interesting questions regarding the mechanisms they use to interact.  We intend
to address the limitations presented hereafter in the near future.

\section*{Limitations} 
Despite achieving SOTA results on open intent classification tasks, our work
has several facets that could be furnished further. Firstly, we tune the last
layer of transformer along with the prefixes, making our method less parameter
efficient than prefixes alone. Other approaches to fine-tuning the last layer
of the transformer during pre-training should be investigated.  Moreover, this
work does not include any other PET method such as adapter tuning \citep{He:21}
or LoRA \citep{Hu:2021}. We anticipate that using other PET methods will reveal
new observations regarding their interaction with partial fine-tuning. We
restrict our study to simple single intent dialogues while industry-deployed
models would likely encounter noise as well as multiple intents. Testing the
robustness of our method under these conditions could be valuable. Lastly, we
plan to research whether our success with prefix-tuning in combination with
partial fine-tuning generalizes to other NLU and financial tasks.

\section*{Ethics Statement}
Recent impressive achievements in NLP thanks to the advent of LLMs do not come
without cost. Most relevant to our paper is the environmental impact and
inequitable distribution of such technologies
\citep{strubell-etal-2019-energy}. The resources required to train a LLM are
large which from the environmental perspective increases our contribution to
climate change and from an equity perspective limits who can access, research,
and use the model. 

While the self-supervised pre-training step often has the greatest resource
requirements, fine-tuning LLMs is undertaken by many more parties following a
model's public release. The numerous task-specific deployments of popular
models likely have greater net CO$_2$ emissions than the initial pre-training.
Our work directly combats this concern by promoting parameter efficient tuning
as an efficacious alternative to relatively expensive fine-tuning. The fraction
of trainable parameters reduces tuning memory requirements, in turn reducing
power consumption and environmental impact. Additionally, the reduction of
required memory enables the adoption of LLMs by those who do not have access to
expensive high-quality hardware or cloud platforms. Finally, storing copies of
the model for each task is efficient. Only a single copy of the frozen LLM is
needed along with the smaller prefixes and in our case, trained last layer of
transformer, resulting in similar benefits as the reduction of memory.

\bibliography{anthology,custom}
\bibliographystyle{acl_natbib}

\appendix

\end{document}